\documentclass[sigconf]{acmart}
\usepackage[utf8]{inputenc}
\usepackage{afterpage}
\setcopyright{none}
\begin{document}
%
\acmConference[Conference'18]{ACM Conference}{July 2018}{Washington, DC, USA}

\title{Do deep reinforcement learning agents model intentions?}

\author{Tambet Matiisen}
\affiliation{%
  \institution{University of Tartu}
}
\email{tambet.matiisen@ut.ee}

\author{Aqeel Labash}
\orcid{0000-0002-2201-2809}
\affiliation{%
  \institution{University of Tartu}
}
\email{aqeel.labash@gmail.com}

\author{Daniel Majoral}
\affiliation{%
  \institution{University of Tartu}
}
\email{daniel.majoral.lopez@ut.ee}

\author{Jaan Aru}
\affiliation{%
  \institution{University of Tartu}
}
\email{jaan.aru@ut.ee}

\author{Raul Vicente}
\affiliation{%
  \institution{University of Tartu}
}
\email{raulvicente@gmail.com}
\begin{abstract}
Inferring other agents' mental states such as their knowledge, beliefs and intentions is thought to be essential for effective interactions with other agents. Recently, multiagent systems trained via deep reinforcement learning have been shown to succeed in solving different tasks, but it remains unclear how each agent modeled or represented other agents in their environment. In this work we test whether deep reinforcement learning agents explicitly represent other agents' intentions (their specific aims or goals) during a task in which the agents had to coordinate the covering of different spots in a 2D environment. In particular, we tracked over time the performance of a linear decoder trained to predict the final goal of all agents from the hidden state of each agent's neural network controller. We observed that the hidden layers of agents represented explicit information about other agents' goals, i.e. the target landmark they ended up covering. We also performed a series of experiments, in which some agents were replaced by others with fixed goals, to test the level of generalization of the trained agents. We noticed that during the training phase the agents developed a differential preference for each goal, which hindered generalization. To alleviate the above problem, we propose simple changes to the MADDPG training algorithm which leads to better generalization against unseen agents. We believe that training protocols promoting more active intention reading mechanisms, e.g. by preventing simple symmetry-breaking solutions, is a promising direction towards achieving a more robust generalization in different cooperative and competitive tasks.
\end{abstract}


\maketitle

\section{Introduction}


The ability of humans to infer the mental states of others such as their beliefs, desires, or intentions is called Theory of Mind (ToM) \cite{apperly2010mindreaders, aru2018deep}. Inferring other agents' intentions gives an advantage both in cooperative tasks, where participants have to coordinate their activities, and competitive tasks, where one might want to guess the next move of the opponent. Predicting other agent's unobservable intentions from a few observable actions has practical applications as for example in self-driving cars where during lane merge the controlling agent has to determine whether the other driver is going to let it in \cite{chater2018negotiating,dinggame}.

\par In this work, we investigate to which degree artificial agents trained with deep reinforcement learning develop the ability to infer the intentions of other agents. Our experiments are based on a cooperative navigation task from \cite{lowe2017multi} where three agents have to cover three landmarks and coordinate between themselves which covers which (see Figure  \ref{fig:task}). We apply a linear readout neuron on each agent’s hidden state and try to predict the final landmark covered by other agents at the episode end. If early in the episode the other agents' final landmark can be accurately predicted, then we could claim that the agents are representing information about others' specific goals, and hence to some extent infer their intentions.

\par In our experiments we indeed show that the intentions of other agents can be decoded from an agent’s hidden state using a linear decoder, while the same information cannot be linearly decoded from the observations (based on which the hidden state is computed). This means that agents apply a learned transformation to the observation which makes this information more explicit in hidden layer 1. Interestingly, the same information can be decoded less accurately from hidden layer 2.

\par 
In our work we also show that training multiple agents jointly using reinforcement learning leads to severe co-adaptation where agents overfit to their training partners and do not generalize well to novel agents. For example, we demonstrate that in a cooperative navigation task the agents trained using multiagent deep deterministic policy gradient (MADDPG) algorithm \cite{lowe2017multi} develop favorite landmarks and these are different for all three trained agents. The lack of generalization is exposed when a MADDPG-trained agent is put together with two “Sheldon” agents - agents that always go to the same fixed landmark\footnote{Named after a character in “The Big Bang Theory” who insisted on always sitting in the same spot.}. We show that performance of an agent degrades substantially when the only remaining available landmark is its least favorite one.

\par Finally we propose simple changes to the MADDPG algorithm that make this problem less severe. In particular, randomizing the order of agents for each episode improves the generalization result while ensembling suggested in \cite{lowe2017multi} does not.

The main contributions of the paper are:
\begin{itemize}
    \item We show that deep reinforcement learning agents trained in cooperative settings learn models which contain explicit information about the intentions of other agents.
    \item We show that jointly trained reinforcement learning agents co-adapt to each other and do not generalize well when deployed with agents that use unseen policies.
    \item We propose simple changes to the MADDPG algorithm that alleviate the above problem to some degree.
\end{itemize}

The code is available at \url{https://github.com/NeuroCSUT/intentions}.

\section{Methods}\label{section.methods}
\afterpage{
\begin{figure*}[ht]
    \centering
    \includegraphics[width=1\textwidth,height=1\textheight,keepaspectratio]{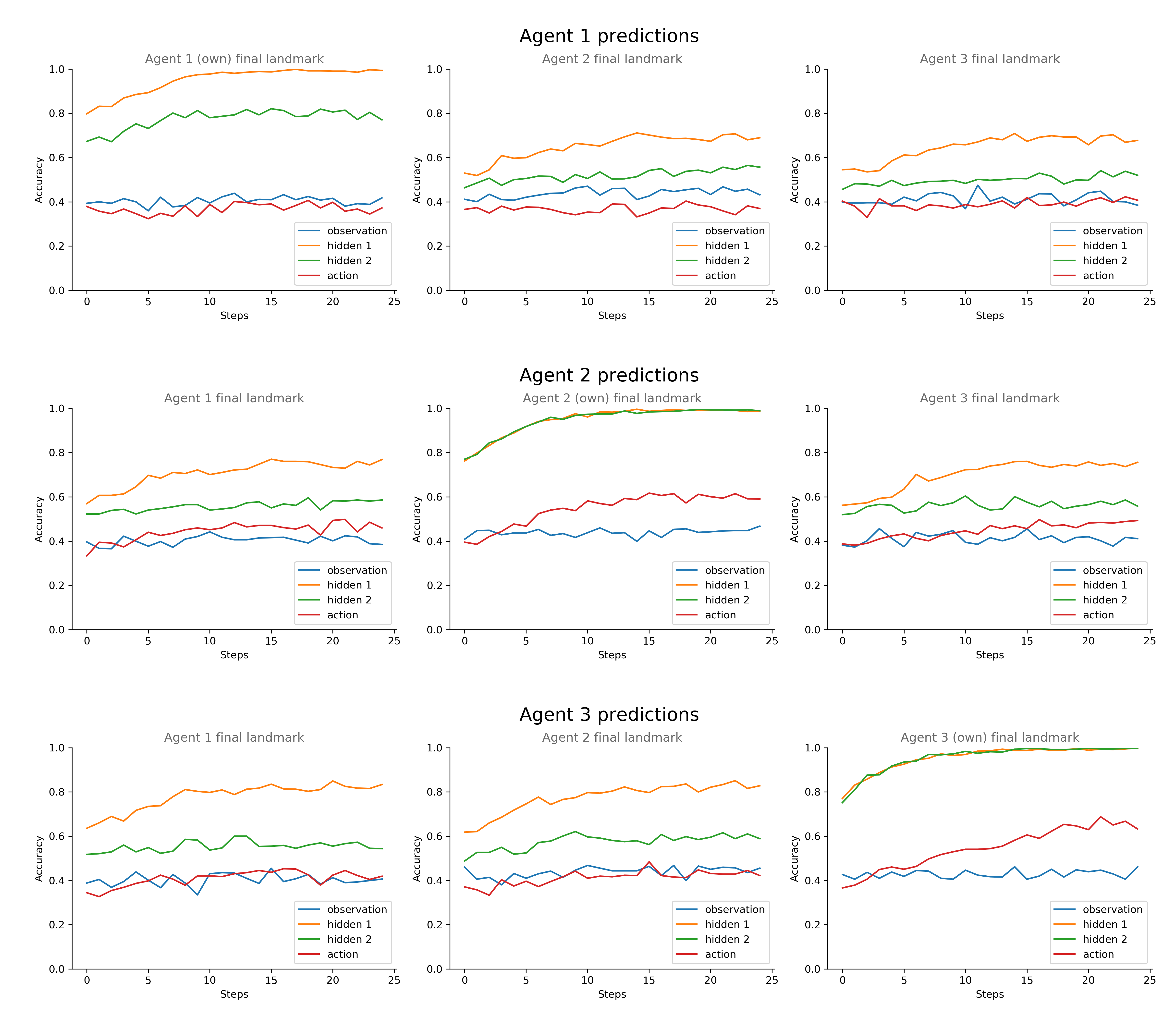}
    \caption{Test set accuracy of a linear read-out neuron for MADDPG + shuffle scheme. All 9 combinations of 3 agents predicting the final landmarks of other 3 agents are shown, including the agent predicting its own final target. Y-axis is the test set accuracy of a linear read-out neuron, X-axis is timestep of an episode.}
    \label{fig:coop_navi_shuffle}
\end{figure*}

}
\begin{figure}
\centering
    \includegraphics[scale=0.25]{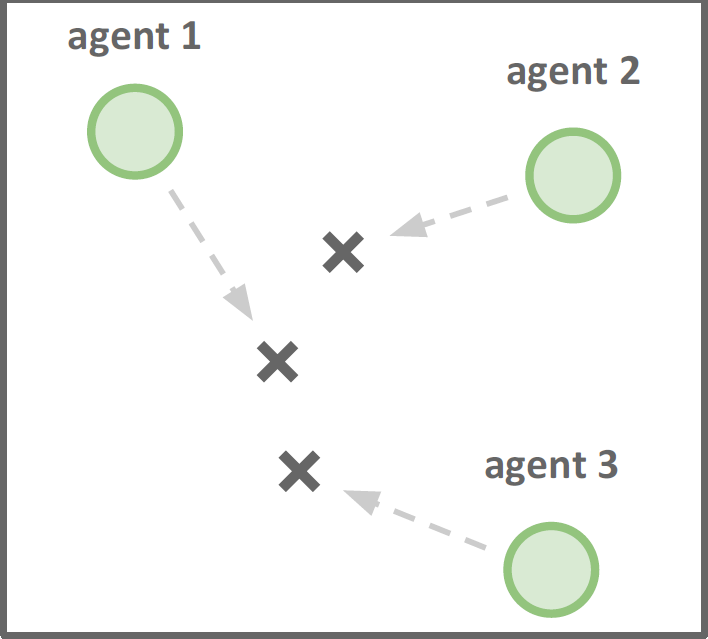}
    \caption{Cooperative navigation task. Circles are agents, crosses are landmarks. The rewarding scheme incentivizes the coverage of each landmark by a different agent without collisions. Image from \cite{lowe2017multi}.}
    \label{fig:task}
\end{figure}

Our experiments are based on a cooperative navigation task described in \cite{lowe2017multi}. In this task three agents try to cover three landmarks and have to coordinate which agent covers which landmark (see Figure \ref{fig:task}). The reward at every time step is 

\begin{equation*}
    r = \sum_{i}min_j(d_{ij})-c
\end{equation*}

\noindent where $d_{ij}$ is the distance from landmark $i$ to agent $j$ and $c$ is the number of collisions. These rewards incentivize that there is exactly one agent close to each landmark and there are as few collisions as possible. The observation of each agent consists of 14 real values: velocity of the agent (2), position of the agent (2), egocentric coordinates of all landmarks (6) and egocentric coordinates of all other agents (4). Action of an agent consists of 5 real values: acceleration in four possible directions (only positive values) and a dummy value for no action.

\par We follow the basic training scheme from \cite{lowe2017multi} and we use the MADDPG algorithm with default settings. The network has two hidden layers with 128 nodes each. We train the agents for 100,000 episodes, when approximately the convergence happens. Thereafter we evaluate the agents for 4,000 episodes and record the observations, hidden states (from both layers) and actions of each agent.

\par To decode other agents' goals we train a linear readout neuron to predict the final landmark (the landmark covered at the final timestep of the episode) of other agents from each agent’s hidden state. This is treated as a classification task with three landmarks as the three classes (the readout model has three outputs). For comparison we also try to predict the final landmark from the observation (network's input) and the actions (network's output). We separate the 4,000 episodes randomly into training and test set (test set size was 25\%) and report classification accuracy on the test set. The training is done separately for each of the 25 time steps (all episodes are of the same length). Use of a linear model guarantees that the information is present explicitly in the model's input.

\par To test the generalization ability of the agents we put them together with two “Sheldon” agents - agents that go straight to a fixed landmark ("their spot"). This leaves one landmark free for the trained agent to cover. We run an evaluation for 9 possible combinations of 3 agents and 3 free landmarks. One evaluation lasts 4000 episodes and we report the percentage of episodes where all landmarks were covered. We compare this with the same measure from previous evaluations where all agents were trained agents. All results are averaged over 5 random seeds.

\par Finally we tried three possible modifications to MADDPG algorithm to alleviate the generalization issues:

\begin{itemize}
    \item \textbf{MADDPG + shuffle:} randomize the order of agents for each episode. The order determines the position of other agents' data in  agent's observation. Randomizing makes it impossible to have fixed assumptions about other agents’ behavior.
    \item \textbf{MADDPG + shared:} use a shared model for all agents, making them basically equivalent and eliminating the option for landmark biases.
    \item \textbf{MADDPG + ensemble:} use an ensemble of agents for each position as suggested in \cite{lowe2017multi}. The agents in an ensemble develop different policies because of different random initialization and different training samples they see from replay. This increases the diversity of partners any agent experiences, which forces it to develop more general strategies.
\end{itemize}


\section{Results}

\subsection{Reading out intentions}
\label{subsection.Reading.out.intentions}

We observed that the agents trained with vanilla MADDPG had preferences for certain landmarks (see also section \ref{subsection.Generalization.gap}) and hence in many cases the readout neuron could achieve a high classification performance by just producing a constant prediction (see Figure \ref{fig:4} in appendix \ref{subsection.Supplementary.Figures}). To remove such class imbalance, we used MADDPG + shuffle scheme for assessing the decoding performance. Figure \ref{fig:coop_navi_shuffle} shows the readout neuron prediction results. We make the following observations:
\begin{itemize}
    \item The agent predicts its own final landmark generally better than others’, which is expected, because it has direct access to the hidden state that guides the actions.
    \item The final landmark of other agents can be predicted numerically better from hidden layer 1 activations than from observations or from hidden layer 2 activations. While all relevant information is already contained within the observation (because hidden states are computed from observations) it is more explicitly represented in the hidden layers and can be successfully decoded with a simple linear model. Representations in hidden layer 2 presumably focus more on the policy (the actions to be chosen) and therefore lose some of the information about intention.
    \item Output of the network (the actions) is completely uninformative for predicting the final landmark of other agents, which is expected. Prediction accuracy from actions coincides with the accuracy from random noise (not shown for brevity) and is close to the chance level (33\%).
    \item The accuracy of landmark prediction from hidden layer 1 is 50-70\%. While not perfectly accurate, it is clearly above chance level. Interestingly, the final landmarks of other agents can be predicted above chance level already in the first timestep. There is also an increase of prediction accuracy as the episode progresses, which is expected.
\end{itemize}

Video showcasing the intention readouts is available here:\\ \url{https://youtu.be/3-pMUPPo970}.

\subsection{Generalization gap}
\label{subsection.Generalization.gap}

As noted before, agents trained with vanilla MADDPG algorithm developed preferences for certain landmarks (see Table \ref{table:1}). In addition to making reading out intentions uninformative due to class imbalance problem, it was also clear that the trained agents had co-adapted to each other and as such would not generalize when put together with agents unseen during training.

\begin{table}[h]
\centering
\caption{Percentage of episodes where a given agent covered a given landmark with vanilla MADDPG algorithm. Only those episodes are counted where all three landmarks were covered by exactly one agent (48\% of all episodes).}
\begin{tabular}{|l|c|c|c|} \hline
&landmark 1&landmark 2&landmark 3\\ \hline

agent 1&74\%&25\%&1\%\\ \hline
agent 2&25\% &75\%&0\%\\ \hline
agent 3&1\%&0\%&99\%\\ \hline

\end{tabular}
\label{table:1}
\end{table}

\par To quantify the lack of generalization we put a trained agent together with two “Sheldon” agents, each of which always goes to a fixed landmark. This test measures the ability of the agent to adapt its policy to an unforeseen situation where only one landmark is free for it to achieve the goal. Indeed the agents showed an inability to adapt to the situation when the free landmark was their least favorite, see Table \ref{table:2} for an example and compare with Table \ref{table:1}.

\begin{table}[h]
\centering
\caption{Percentage of episodes where agents were able to cover all landmarks, when two ``Sheldon'' agents covered two landmarks and given agent had only one free landmark to cover. When paired with two trained agents the reference number to compare with is 48\%.}
\begin{tabular}{|l|c|c|c|} \hline
&landmark 1&landmark 2&landmark 3\\ \hline

agent 1&65\%&42\%&3\%\\ \hline
agent 2&28\% &75\%&2\%\\ \hline
agent 3&5\%&2\%&78\%\\ \hline
\end{tabular}
\label{table:2}
\end{table}

While Table \ref{table:1} and Table \ref{table:2} showcase a single training run for clarity, similar pattern occurred in all runs.

\subsection{Improving generalization}

\begin{figure}
    \centering
    \includegraphics[width=0.45\textwidth]{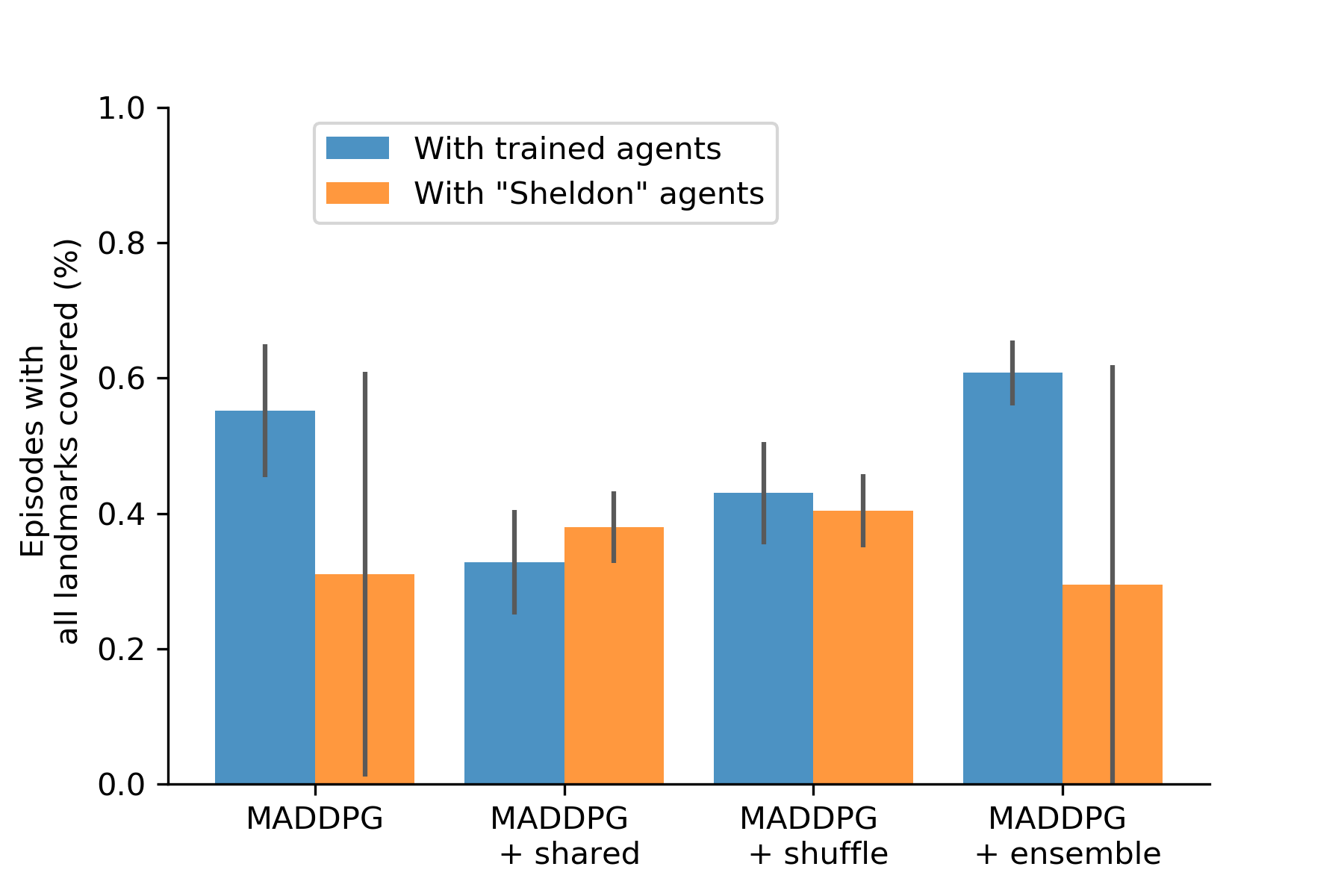}
    \caption{The percentage of episodes where all three landmarks were covered for different training schemes. The blue bars represent the means and standard deviations over 5 runs across all trained agents. The orange bars represent the means and standard deviations across $5\times9$ runs with two ``Sheldon'' agents and one trained agent.}
    \label{fig:3}
\end{figure}

To improve the generalization of MADDPG algorithm we tried out three different modifications, as described in section \ref{section.methods}. To test the generalization ability we report the percentage of episodes where the agents covered all three landmarks. This number is averaged over 5 training runs and over 9 agent-landmark combinations as seen in Table \ref{table:2}. Figure \ref{fig:3} shows the results.

We make the following observations:
\begin{itemize}
    \item \textbf{Vanilla MADDPG} agents achieve very good results when evaluated against other trained agents, but fail when confronted with agents with unseen policies. Large standard deviation on orange bar results from the fact that for favorite landmarks the success rate is very high and for the least favorite landmarks the success rate is very low (see Table \ref{table:2} for an example).
    \item \textbf{MADDPG + shared} scheme was much worse than Vanilla MADDPG when evaluated against other trained agents, but surprisingly achieved better generalization.
    \item \textbf{MADDPG + shuffle} scheme achieves consistent success rate both with trained agents and “Sheldon” agents.
    \item \textbf{MADDPG + ensemble} scheme improves the result with trained agents but does not fix the generalization gap.
\end{itemize}

Video showcasing the generalization improvements is available here: \url{https://youtu.be/r5jMpdC_pSk}.

\section{Discussion}

Inferring the intentions of others is the basis for effective cooperation \cite{tomasello2010origins}. We studied a very simplified version of cooperation where three agents needed to cover three landmarks. We observed that the agents learn to model the intentions of other agents: already in the beginning of the episode the activity of the hidden layers of a particular agent carried information about the other agents’ targets. While one could make the argument that our result is a far cry of reading the intentions of others, we would contend that this aspect of intention modeling is a necessary building block of reading more complex intentions. Furthermore, studying human level intentions (and ToM) has proven to be complicated \cite{siegal2008marvelous,call2008does,apperly2010mindreaders,heyes2014cultural}; studying simple tasks with agents whose representations can be opened will provide unique insights into the emergence of more complex aspects of ToM \cite{aru2018deep}. In particular, being able to manipulate the network architecture and the components of the system allows one to answer which aspects really matter for solving a particular task. For example, inferring and generalizing intentions seems to require explicit memory (i.e. knowledge about the behavioral patterns of specific other agents), but our current work shows that rudimentary intention reading can be done even without a specific memory store. It might be possible that for more complex scenarios recurrent neural networks or networks with external memory are needed. 

\subsection{Generalization in multiagent setups}

The lack of generalization in reinforcement learning has been criticized before \cite{machado2018revisiting}. Especially problematic is the fact that we tend to test our agents in the same environment where they were trained in. Multiagent training adds another dimension to the generalization problem - the agents should also perform well against opponents and with partners unseen during training.

In the present work we observed that when trained with the same partners, agents overfit to the behavior of their partners and cannot cooperate with a novel agent. While the issue of co-adaptation during training was also raised in \cite{lowe2017multi}, they mainly pointed it out in the context of competitive tasks. We show that it is just as much of a problem in cooperative tasks. Furthermore, we demonstrated that the ensembling approach suggested in \cite{lowe2017multi} does not fix it. More thorough analysis of the overfitting problem is presented in \cite{lanctot2017unified}.

\par In principle the solution is simple - the agents need to experience a variety of partners during training to generalize well. Similar to data augmentation used in supervised training we would need to ``augment'' our policies in various ways to produce the widest variety of training partners. Unfortunately it is not clear how to achieve this in an automated and generalizable way. One proposed approach is to learn maximum entropy policies as done in \cite{eysenbach2018diversity}. However, it is not clear if that process would ever produce ``Sheldon'' policies used in our experiments, or if they are actually needed.

\par Our MADDPG + shuffle scheme randomizes the order of agents at the beginning of every episode. When the policy is stateless (as it is in our case) this could also be done at every time step. One might be concerned that this deviates from the on-policy learning regime. Our reasoning is that sampling between three policies could be considered as a (stochastic) super-policy that consists of three sub-policies. For the super-policy the learning would still be on-policy. Nevertheless, in our experiments we only used episode level shuffling because it also generalizes to policies with memory (i.e. recurrent neural networks) and is conceptually easier. In practice we did not observe much difference.



\section{Conclusions}

When the controlling agent of a self-driving car performs a lane change, it needs to cope with any kind of behavior from other drivers, however hostile or incompetent they are. Inferring the intentions of other agents is therefore crucial to behave in a reliable manner. For example, the controlling agent of a self-driving car cannot expect all other cars to run the same build of the same software, or even, any software at all. So they need to generalize to unforeseen situations and behaviors of other drivers.

\par In this work we showed that deep reinforcement learning agents indeed learn to model intentions of other agents when trying to solve a cooperative task. In cooperative navigation task the final target of another agent could be predicted better from the hidden layer activations than from the observation. Because the hidden layer is computed from observation, the transformation applied by agent must make this information more explicit, so that it can be read out better with a linear neuron. This also confirms that linear read-out neurons are a good technique for examining the learned network.

\par Trying to read out the intentions of agents exposed the lack of generalization in learned models. Trained agents co-adapted to specific behaviors of each other and failed consistently when put together with unseen agents. We showed that simple shuffling of all agents at each episode improves the generalization while ensembling does not. While this alleviates the generalization gap somewhat, a more robust solution would be to train the agents against opponents using a diverse range of policies. More generally, training paradigms promoting an active inference of other agents' intentions and specific goals is a promising road to narrow the generalization gap. 

\section*{Acknowledgments}
The authors thank the financial support from The Estonian Research Council through the personal research grant PUT1476. We also acknowledge funding by the European Regional Development Fund through the Estonian Center of Excellence in in IT, EXCITE.

\bibliographystyle{abbrv}
\bibliography{acmart}

\appendix
\onecolumn
\section{Supplementary Figures}
\label{subsection.Supplementary.Figures}

\begin{figure*}[h]
    \centering
    \includegraphics[width=1\textwidth,clip]{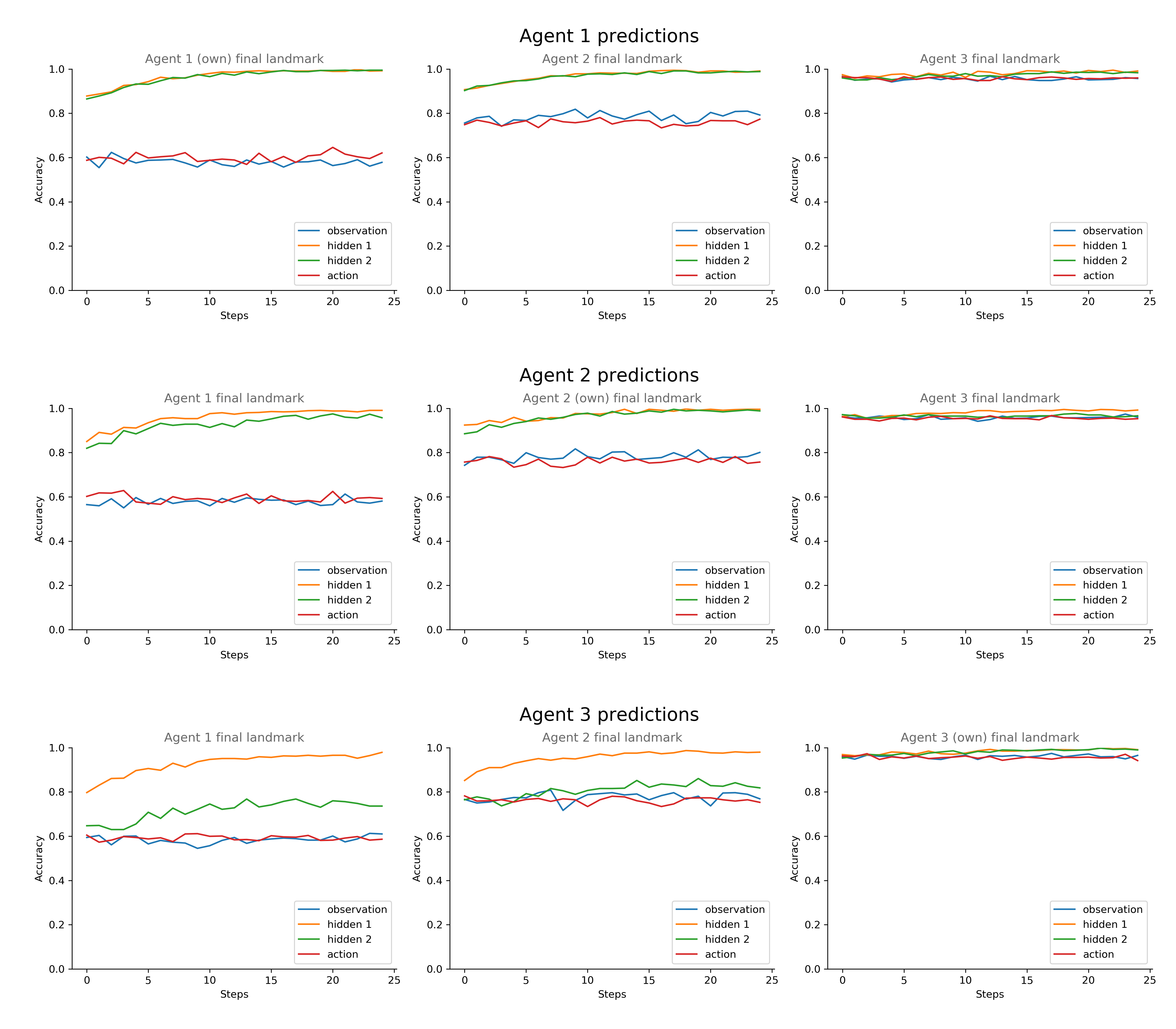}
    \caption{Test set accuracy of linear read-out neuron for vanilla MADDPG algorithm. The predictions are unusually good because of biases of learned agents - the read-out model can score well by just producing constant prediction. For example agent 3 can be predicted perfectly because in 99\% of the cases it goes to landmark 3 (see Table \ref{table:1}). Still the comments from section \ref{subsection.Reading.out.intentions} hold.}
    \label{fig:4}
\end{figure*}
\twocolumn
\end{document}